%% file: main.tex
\pdfoutput=1
\documentclass[lettersize,journal]{IEEEtran}
\usepackage{amsmath,amsfonts, amssymb}
\usepackage{algorithmic}
\usepackage{array}
\usepackage{textcomp}
\usepackage{color}
\usepackage{stfloats}
\usepackage{url}
\usepackage{verbatim}
\usepackage{graphicx}
\usepackage{duckuments}
\usepackage{bm}
\usepackage{rotating}
\usepackage{subfigure}
\usepackage{comment}
\usepackage{tabularx}
\usepackage{colortbl}
\usepackage{hhline}
\usepackage{multirow}
\usepackage{cite}
\usepackage{siunitx}
\usepackage[ruled, lined, linesnumbered, commentsnumbered, longend]{algorithm2e}
\usepackage[capitalise,noabbrev]{cleveref}
\def\BibTeX{{\rm B\kern-.05em{\sc i\kern-.025em b}\kern-.08em
    T\kern-.1667em\lower.7ex\hbox{E}\kern-.125emX}}
\usepackage{balance}

\begin{document}
\title{Generating Whole-Body Avoidance Motion through Localized Proximity Sensing}
\author{Simone Borelli, Francesco Giovinazzo, Francesco Grella*, Giorgio Cannata
\thanks{All the authors are affiliated with the Department of Informatics, Bioengineering, Robotics and Systems Engineering (DIBRIS), Università di Genova, Via all'Opera Pia 13, 16145 Genova, Italy}
\thanks{$^{*}$Corresponding author e-mail: francesco.grella@edu.unige.it}
\thanks{This work was supported by the SESTOSENSO (HORIZON EUROPE Research and Innovation Actions under GA number 101070310).}}

\maketitle

\begin{abstract} 
This paper presents a novel control algorithm for robotic manipulators in unstructured environments using proximity sensors partially distributed on the platform. The proposed approach exploits arrays of multi-zone Time-of-Flight (ToF) sensors to generate a sparse point cloud representation of the robot's surroundings. By employing computational geometry techniques, we fuse the knowledge of robot's geometric model with ToFs sensory feedback to generate whole-body motion tasks, allowing to move both sensorized and non-sensorized links in response to unpredictable events such as human motion. In particular, the proposed algorithm computes the pair of closest points between the environment cloud and the robot's links, generating a dynamic avoidance motion that is implemented as the highest priority task in a two-level hierarchical architecture. Such a design choice allows the robot to work safely alongside humans even without a complete sensorization over the whole surface. Experimental validation demonstrates the algorithm's effectiveness both in static and dynamic scenarios, achieving comparable performances with respect to well established control techniques that aim to move the sensors mounting positions on the robot body. The presented algorithm exploits any arbitrary point on the robot's surface to perform avoidance motion, showing improvements in the distance margin up to 100 mm, due to the rendering of virtual avoidance tasks on non-sensorized links.

\end{abstract}


\section{Introduction}
\input{01_Introduction/introduction}

\section{Methodology}
\input{02_Methodology/methodology}

\section{Experimental Validation}
\input{03_Experimental_Validation/experiments}

\section{Results and Discussion}
\input{04_Results_and_Discussion/discussion}

\section{Conclusions}
\input{05_Conclusions/conclusions}

{\small
	\bibliographystyle{IEEEtran}
	\bibliography{ref}
}

\section{Biography}
\vspace{11pt}
\vspace{-25pt}
\begin{IEEEbiography}[{\includegraphics[width=1in,height=1.25in,clip,keepaspectratio]{example-image-a}}]{Simone Borelli}
received the bachelor’s degree in computer engineering from the University of Genova, Genoa, Italy, in 2022.
He is a Master student in robotic engineering and is currently working on his thesis project at the Mechatronics and Automatic Control Laboratory (MACLab) in University of Genoa, Italy.  His research interests include robotics, robot sensing and control systems.
\end{IEEEbiography}
\vspace{-25pt}
\begin{IEEEbiography}[{\includegraphics[width=1in,height=1.25in,clip,keepaspectratio]{example-image-b}}]{Francesco Giovinazzo} is a third year Ph.D. student at the Mechatronics and Automatic Control Laboratory (MACLab) in University of Genoa, Italy. In 2021 his Master thesis project was awarded as 'Best thesis of 2021' from the 'Comitato Elettrotecnico Italiano (CEI)'. He is currently working on reactive robot control based on distributed tactile and proximity sensors for safe Human-Robot Interaction.
\end{IEEEbiography}
\vspace{-25pt}
\begin{IEEEbiography}[{\includegraphics[width=1in,height=1.25in,clip,keepaspectratio]{example-image-c}}]{Francesco Grella} (Member, IEEE)
received the bachelor’s
degree in biomedical engineering and the master’s degree in robotics engineering from the
University of Genova, Genoa, Italy, in 2017
and 2019, respectively. He pursued the
Ph.D. degree in robotics and autonomous systems with the Università degli Studi di Genova,
Genoa. He is a Post-Doctoral research rellow at the Università degli
Studi di Genova. His research interests focus on
sensor-based robot control, tactile-based robotic
perception, and tactile sensor design and fabrication.
\end{IEEEbiography}
\vspace{-25pt}
\begin{IEEEbiography}[{\includegraphics[width=1in,height=1.25in,clip,keepaspectratio]{example-image-a}}]{Giorgio Cannata} (Member, IEEE) received the Laurea degree in electronic engineering from the University of Genoa, Genoa, Italy, in 1988.
He is a Full Professor of Automatic Control at the Polytechnic School of Engineering, University of Genoa. He is the Scientific Coordinator of the Mechatronics and Automatic Control Laboratory (MACLab), Department of Informatics, Bioengineering, Robotics and System Engineering (DIBRIS), University of Genoa. He is coordinator of the National Ph.D. program in Robotics and Intelligent Machines (DRIM). His current research interests include robotics, mechatronics, and robot sensing and control systems
\end{IEEEbiography}

\end{document}

%% file: 01_Introduction/introduction.tex
Robotic manipulation in unstructured environments represents a significant challenge since robots must operate safely and efficiently in the presence of dynamic and unpredictable obstacles. Traditional control approaches often rely on external sensing architectures based on cameras, tactile sensors, and more recently proximity sensing. 

Visual-servoing architectures are a well-established approach relying on image processing techniques to perceive the surrounding space \cite{kim_moving_2012, lagisetty_object_2013, scimmi_implementing_2019, Nascimento2020, liu_collision-free_2021}. External vision-based systems, including cameras and motion capture setups, are used to enhance robot perception by monitoring the collaborative workspace, thus ensuring safety in Human-Robot Interaction (HRI) applications \cite{zanchettinPredictionHumanActivity2019, bascettaSafeHumanrobotInteraction2011, tellaecheHumanRobotInteraction2015, duMarkerlessHumanRobot2015}. However, these solutions rely on well-calibrated, structured environments and are prone to failure in cluttered scenarios due to limitations in the field of view and sensitivity to light conditions. 

\begin{figure}[t!]
    \centering
    \includegraphics[width=0.48\textwidth]{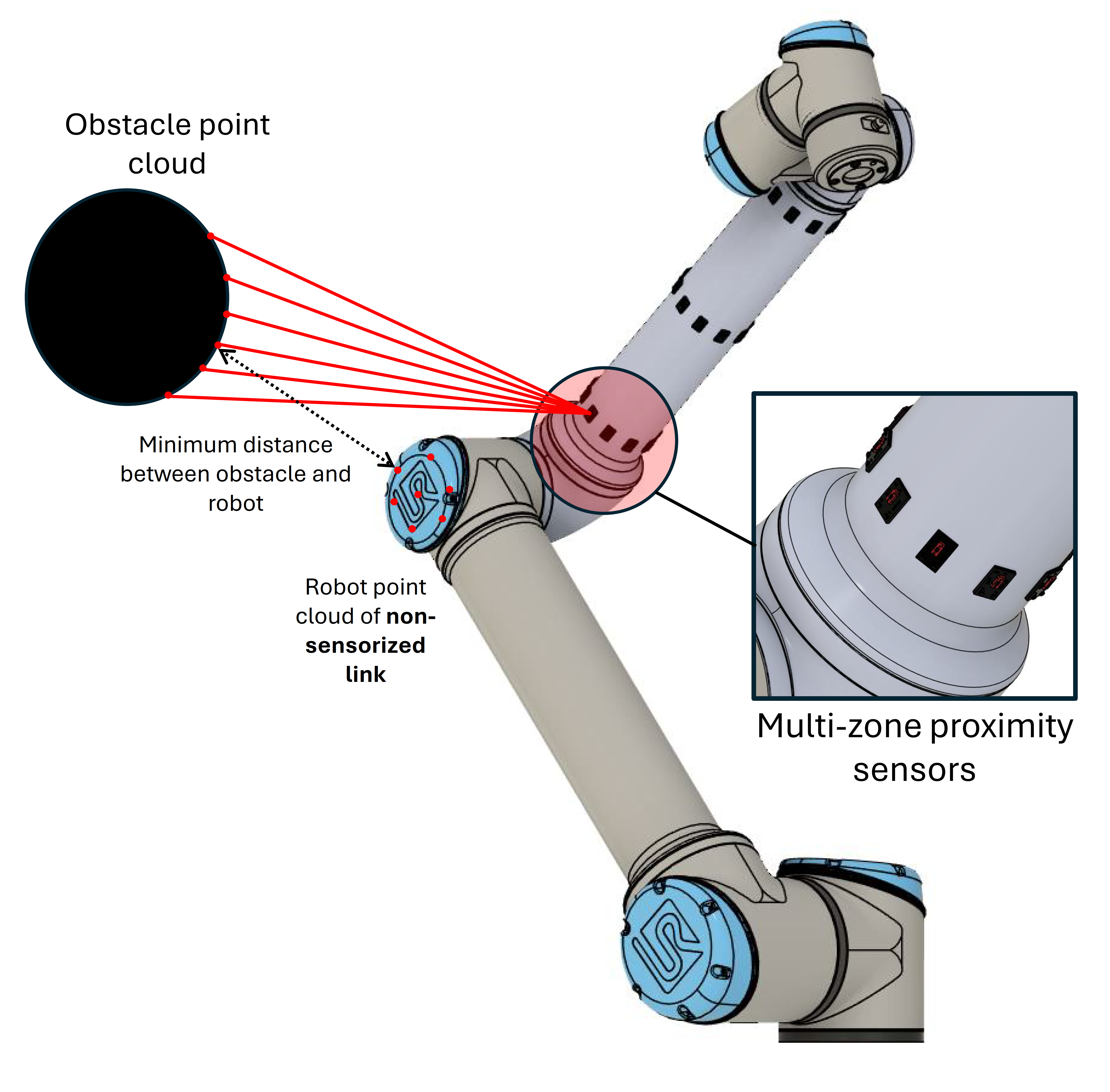}
    \caption{3D model of a robotic manipulator partially covered with multi-zone proximity sensors. On the left side of the picture is highlighted an obstacle detected by one of the sensors. Despite being closer to the second, non-sensorized link of the robot, a traditional approach would use the position of the closest proximity sensor as control point to drive the obstacle avoidance behaviour. Our solution exploits knowledge of the robot's geometry to define a dynamic avoidance task that acts on an arbitrary point over the robot's surface, regardless of its sensorization.   }
    \label{fig:overview}
\end{figure}

Tactile perception is a preferred resource in tasks implying physical contact with the environment, such as autonomous clutter exploration \cite{albiniExploitingDistributedTactile2021, Ye_2022_tactile, Brouwer_2024_tactile, Jiang_2024_tactile} and physical Human-Robot Interaction (pHRI) \cite{grellaVoluntaryInteractionDetection2022}. Both applications require the robotic system to be extensively sensorized all over its surface, leading to so-called \textit{large-area sensing} \cite{cheng_main, TacSuit, maiolinoFlexibleRobustLarge2013, yangRoboticSkinMimicking2024}. In particular, large-area tactile sensing enables the integration of pressure sensors on complex curved surfaces, such as manipulator links, offering capabilities like multi-contact detection and contact geometry reconstruction. Hence motion control algorithms that exploit distributed contact feedback allow robots to safely interact with the environment and gather information on its physical properties. Moreover, tactile sensing is insensitive to variations in light conditions and an extensive coverage with few blind spots allows to perceive contacts coming from any direction. Relying solely on touch when handling multiple sources of uncertainty is impractical and may result in potentially hazardous situations. For instance, in densely crowded environments or safety-critical applications like Human-Robot Collaboration (HRC), tactile perception only provides local awareness of obstacles and cannot detect anything that is not already in contact with the robot. To overcome this limitation, the research community is extensively working on proximity sensing, both as a unique sensing modality \cite{tsujiProximitySkinSensor2019a, Ding_2020_proximity, Ferat_2019} or as part of multi-modal architectures  \cite{hughesRoboticSkinCollision2018, abahMultimodalSensorArray2019, cheng_main, Giovinazzo,  heoProximityPerceptionBasedGrasping2024a}.
Handling human presence within the robot's operational space is a well-established research topic which led to the definition of different approaches. Recent works described in \cite{liOptimalMotionPlanning2024} and \cite{liuIntegratingUncertaintyAwareHuman2024} tackle the problem of safe HRI by adopting uncertainty-aware planning algorithms to estimate human motion. However, the approach described in \cite{liuIntegratingUncertaintyAwareHuman2024} relies on a fixed perception setup based on Vicon cameras, therefore bringing many limitations of camera-based approaches. In \cite{liOptimalMotionPlanning2024} the authors define a task-dependent 'danger-zone' to trigger avoidance behavior and safely accomplish the task. 
In this paper, we propose an alternative approach based on large-area proximity sensors embodied on the manipulator, which becomes independent from any external sensing device, to provide reactive avoidance independently from the task nature and physical implementation.
Range-based approaches relying on LiDAR sensors have been recently presented in the context of human-crowded scenarios as depicted in \cite{kudoLiDARBasedPedestrianFlow2024a}. LiDARs are a preferred choice when dealing with mobile robots, but can represent a limitation in collaborative manipulation tasks due to its fixed mounting and consequent occlusions. 

Proximity-based whole-body control enables the development of obstacle avoidance behaviors that can react to unexpected collisions without the limitations of camera-based algorithms. However, equipping the entire robot body with proximity sensors can be an expensive and complex engineering challenge. As a result, many solutions focus on integrating distributed sensors only on the final few links of the kinematic chain, as these are more prone to collisions \cite{tsujiProximitySkinSensor2019a, Navarro_2016, caroleo2024}.

Proximity servoing algorithms usually rely on distance measurements from sensor mounting locations to generate reactive collision avoidance tasks. However, these algorithms are often based on a limited set of control points -- specifically, the sensor locations on the robot's body -- to move the robot away from detected obstacles \cite{Avanzini2014, Ding_2020_proximity, caroleo2024}. This approach has significant drawbacks in terms of flexibility, robustness, and fault tolerance.

Key limitations include that: \textit{I)} the robot's manipulator body must be fully covered with proximity sensors to avoid blind spots; \textit{II)} in certain configurations, non-sensorized parts of the robot may be closer to the obstacle. As a result, while trying to avoid a collision with a sensorized link, the robot might unintentionally collide with a non-sensorized link;  \textit{III)} if any sensor malfunction occurs, the obstacle avoidance behavior could become unreliable.

To address these limitations, this paper presents a novel control algorithm that uses localized proximity sensing to enable reactive motion behaviors across the entire surface of the robot, including non-sensorized links. The algorithm employs a limited number of multi-zone proximity sensors on a single robot link to detect nearby obstacles. By processing this sparse sensor data using computational geometry techniques, it calculates the minimum distance between the environment and the robot's geometric model. This approach extends reactive control to any point on the robot's body by generating virtual control inputs that are fed to a two-layer task priority architecture \cite{albiniHumanHandRecognition2017}.
In particular, the obstacle avoidance behaviour is given the highest priority, whereas the goal-reaching tasks are performed within the null space of the primary task. This design choice ensures an effective robot operation in complex, crowded and unstructured environments without affecting the safety-related objectives.

 To the best of our knowledge, this is the first work addressing whole-body obstacle avoidance with partial proximity sensorization. Moreover, providing non-sensorized links with reactive avoidance capabilites represents a considerable improvement in terms of sensory integration efforts and costs.

This paper presents the following contributions:
\begin{itemize}
    \item a proximity data processing pipeline that efficiently filters out the robot model from the measurements of the proximity sensors;
    \item an algorithm that extends sensor-based reactive control to the whole manipulator's surface without the need for a complete sensorization;
    \item an experimental evaluation of the proposed approach that considers static obstacles and human-robot interaction;
\end{itemize} 

The remainder of this paper is organized as follows: Section \ref{sec:methodology} details the proposed methodology, focusing on the data processing algorithms and the implemented task priority control architecture. Section \ref{sec:experiments} describes the experiments carried out to validate our approach. Section \ref{sec:discussion} provides the results obtained and discusses the implications and limitations of our algorithm. Finally Section \ref{sec:conclusions} draws the conclusions of this work and provides some insights for future research developments.

\begin{figure}[t]
    \centering
    \includegraphics[width=0.5\textwidth]{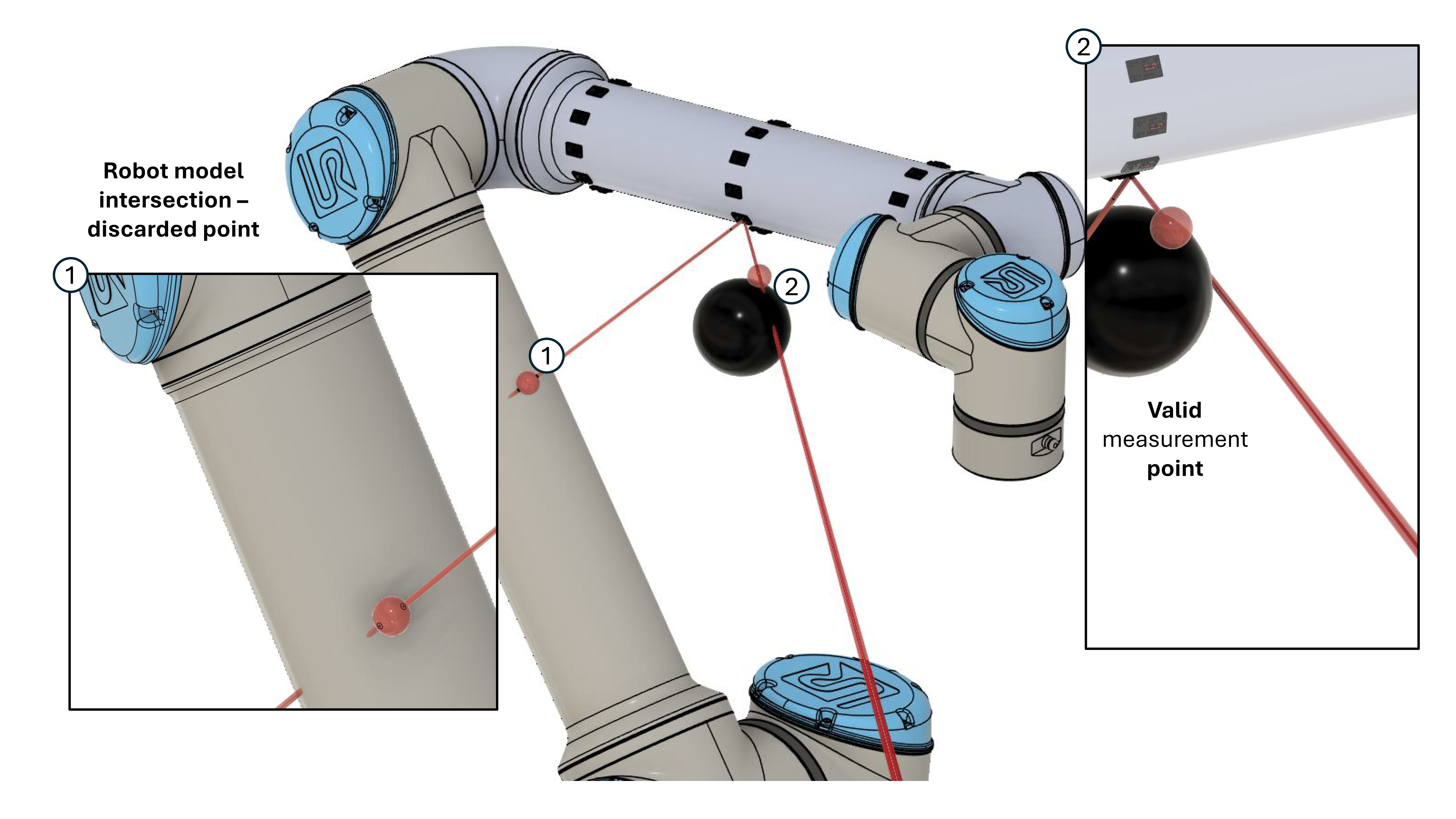}
    \caption{Simplified 3D description of the mesh removal algorithm. A ray is traced between a proximity sensor and all its sampled measurements, expressed as 3D points. If the ray intersects the robot's model, the measurement is compared with the intersection point and if the distance is lower than a specified threshold, the measurement is discarded (1). The point (2) instead is an observation of an obstacle, and the corresponding ray does not intersect with the robot model, therefore the measurement is valid.}
    \label{fig:robot_model_filter}
\end{figure}

%% file: 02_Methodology/methodology.tex
\label{sec:methodology}

The algorithm presented in this paper relies on distance measurement, geometric information about the environment and the robot model to efficiently handle and react to approaching obstacles. However, a set of pre-processing operations must be performed on the raw data in order to ensure a proper identification of measurement corresponding to the robot body, to the static environment and to the objects close to the robotic platform. 

\subsection{Robot Model Filtering}
Multiple proximity sensors are integrated all over the robot surface, as shown in Figure \ref{fig:overview}. In specific joint configurations, ToF sensors Field of View (FoV) might intersect other links of the robotic platform, that would be treated as obstacles 
unless filtered out from the raw point cloud. To overcome such problems we developed an algorithm that processes the raw point cloud, eliminating all measurements corresponding to the robot body.

In the proposed solution we assume to have $N$ proximity sensors distributed on the second link of a robot manipulator, in the known locations $\mathbf{x}_i$, with $i = \lbrace 0, 1, \dots, N \rbrace$ as depicted in  \cref{fig:overview}. 
Each sensor has a squared field-of-view and provides a set of measurements $\mathbf{d} = \lbrace d_1, d_2, \dots, d_M \rbrace$ representing the absolute distance from the object with respect to the measuring device.

Distance measurements can be converted into a small point cloud $P_i$, according to the sensor detection model provided in \cite{vl53l8cx_documentation} and all the collected information can be referred to a common reference frame.

We also assume the full knowledge of the robot model in a form of mesh composed of vertices and faces. In this respect we define $\mathcal{M}(\mathbf{q})$ the mesh where the vertex positions is computed depending on the joint configuration $\mathbf{q}$ with respect to the same reference frame used for each $P_i$.

Then, for each  point $p \in P_i$ we check the intersection with the robot mesh $\mathcal{M}_i$.
In particular, we compute a ray starting from the origin of the measurement $\mathbf{x}_i$ and passing through $p$. 
Then, to check for the intersection we leverage AABB Tree \cite{bibid}, a data structure allowing to perform intersection queries in logarithmic time. Therefore, an AABB tree  $\mathcal{T}(\mathbf{q})$ is computed from the mesh $\mathcal{M}(\mathbf{q})$ at the given robot posture. For each ray, we then check the intersection point between the ray and $\mathcal{T}(\mathbf{q})$.

In case of intersection, it is still necessary to check if the measurement point belongs to the robot mesh. Indeed the ray intersecting the robot model might also pass through an obstacle interposed between the sensor and the robotic platform, as depicted in Figure \ref{fig:overview}. In this case, we compute the distance between $p$ and the intersection point with the robot model and we compare it with a threshold $\epsilon$. 

\begin{figure*}[t]
	\centering
	\subfigure[]{\includegraphics[scale=0.15]{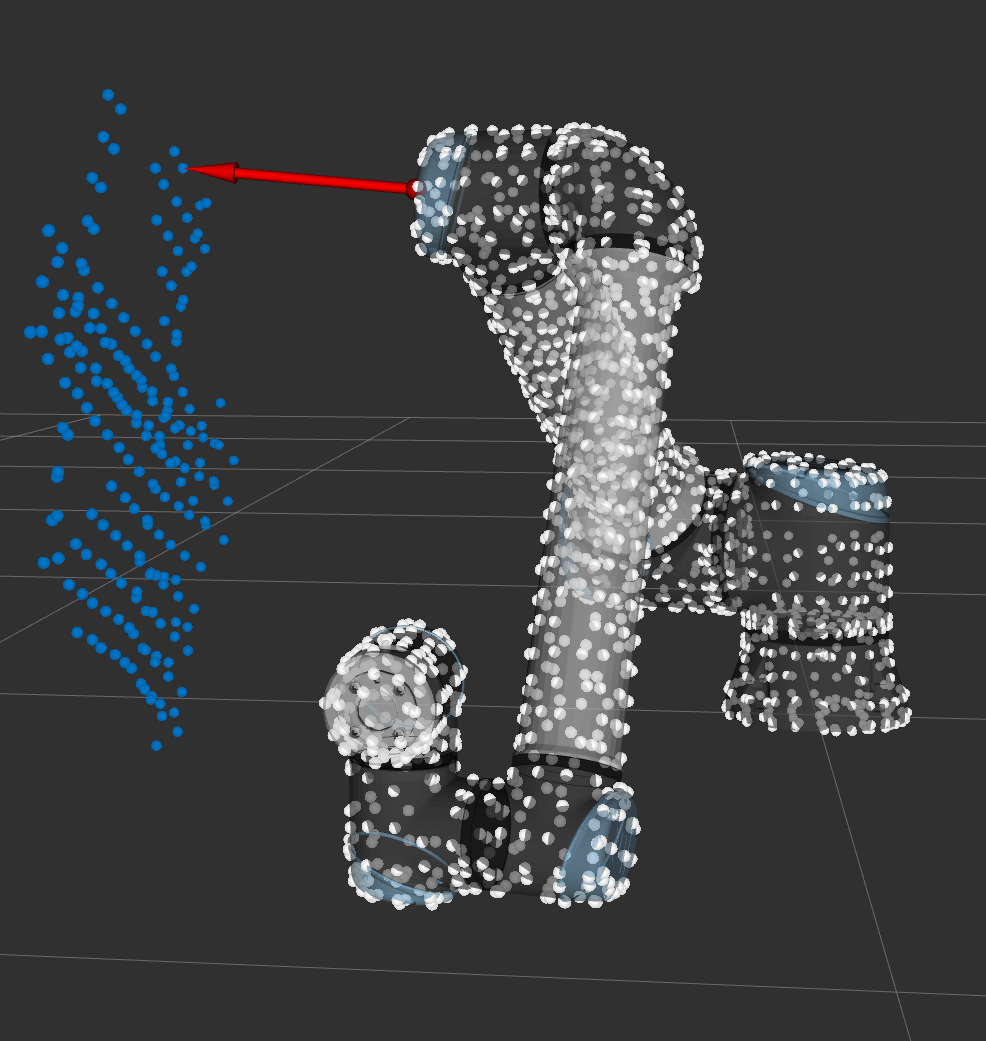}}
        \subfigure[]{\includegraphics[scale=0.15]{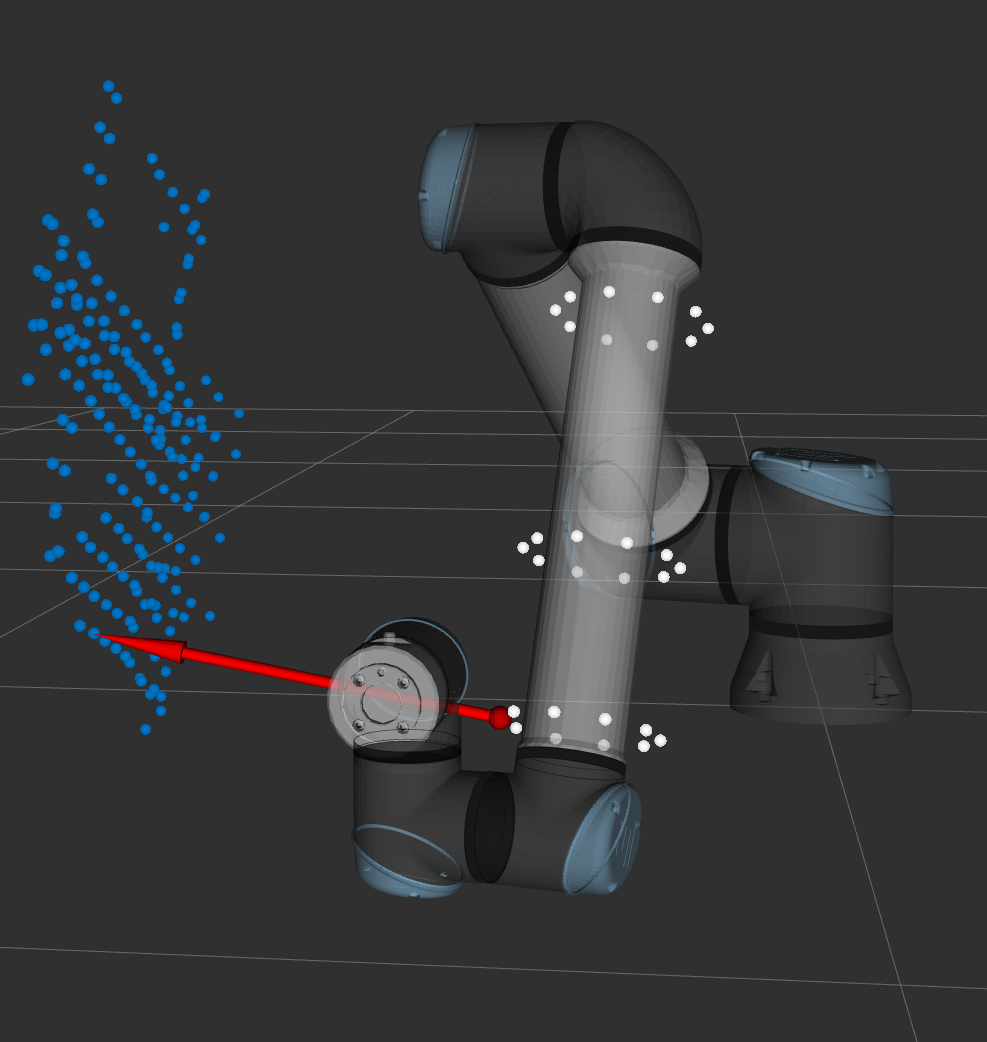}}
        \subfigure[]{\includegraphics[scale=0.1522]{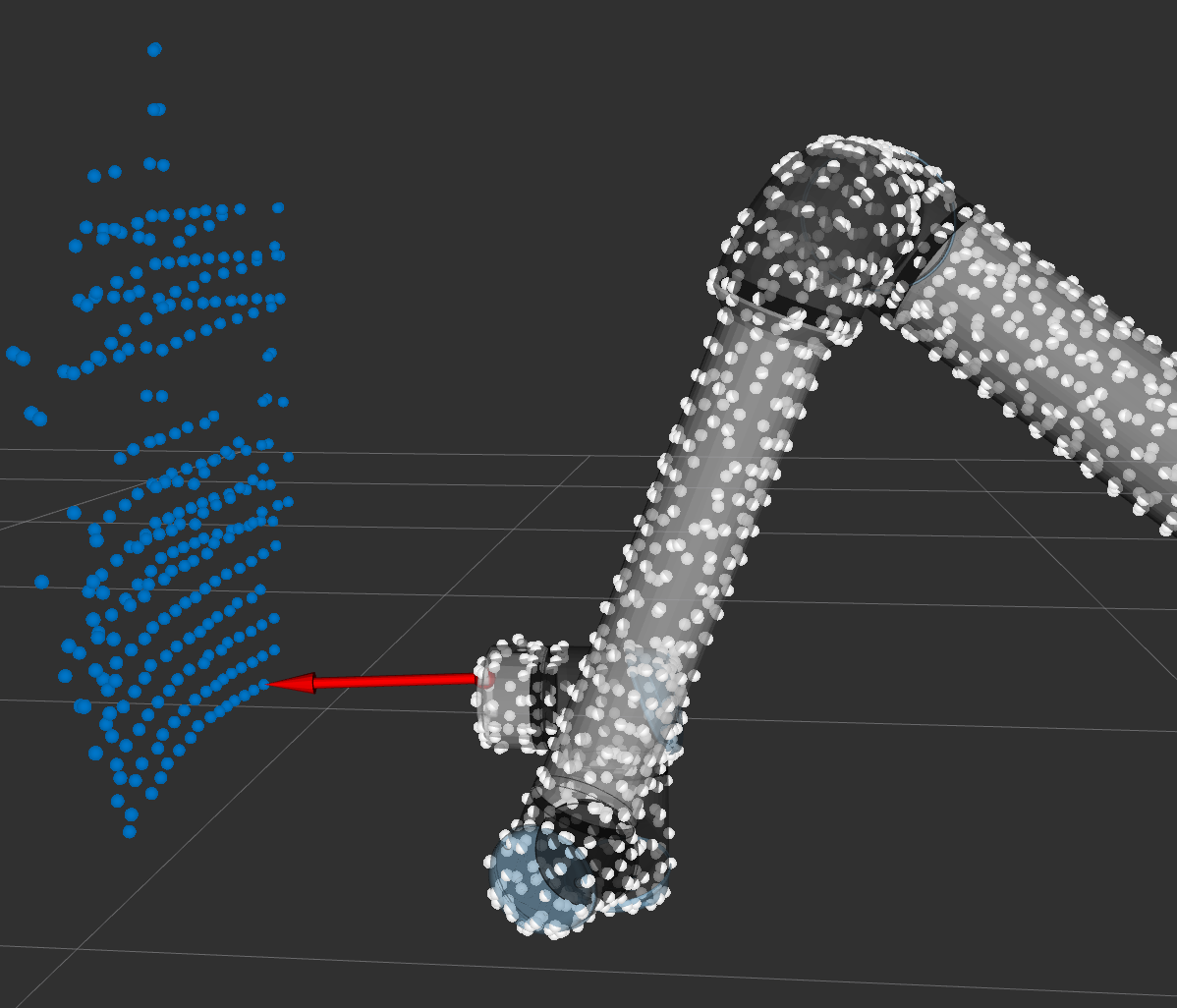}}
	\subfigure[]{\includegraphics[scale=0.15]{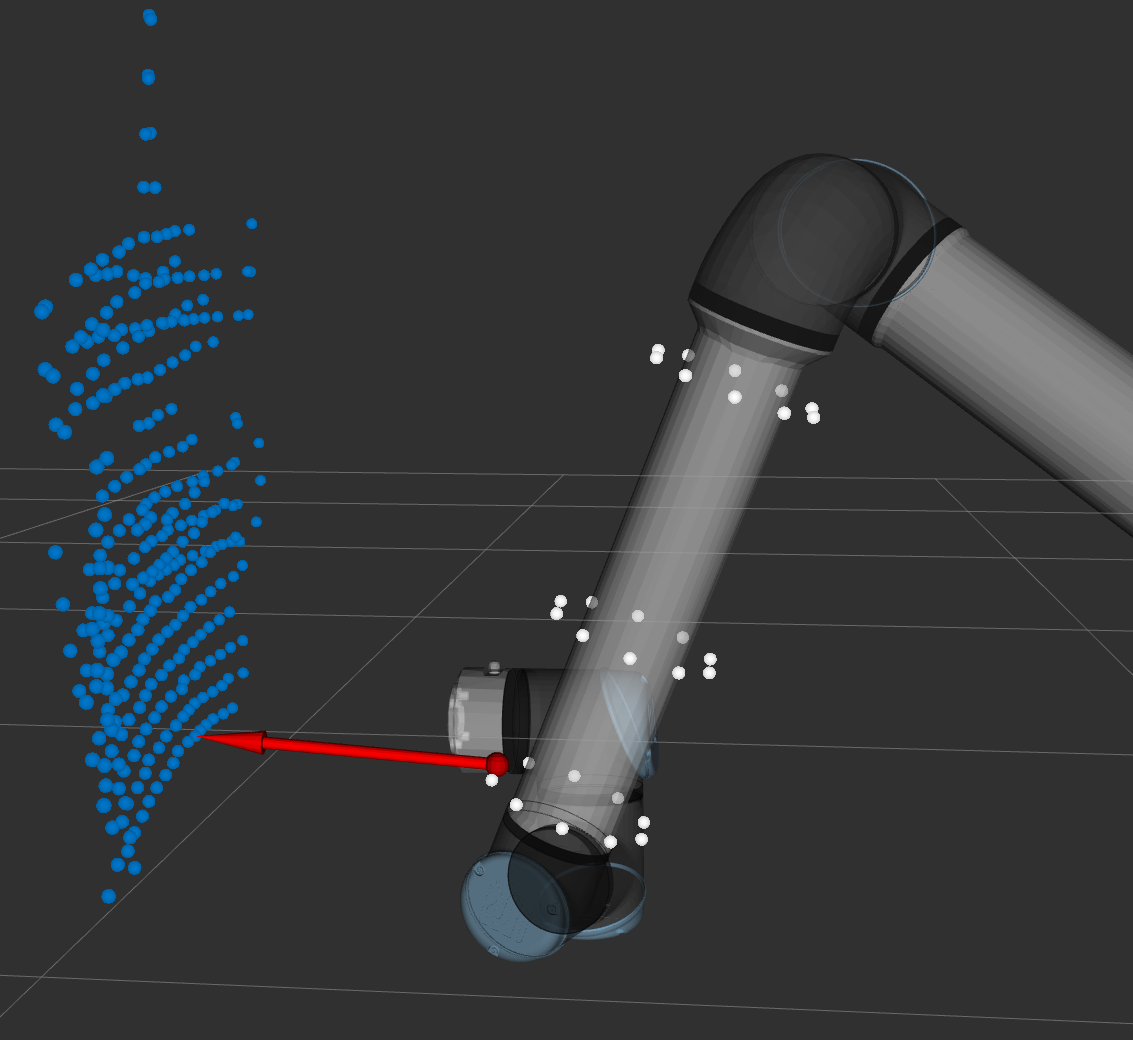}}
	\caption{Visual comparison of the whole-body minimum distance pair algorithm and its counterpart based on sensors mounting points. The white point cloud represents a downsampled model of the robot computed by its visual meshes (a) and (c) and proximity sensors locations in (b) and (d). The blue cloud represents an obstacle resembling our proposed experimental configurations. The red arrow is the vector that has the pair of closest points as its origin and tip.}
	\label{fig:mindist_methods}
\end{figure*}

The full algorithm is showed in Alg. 1.

	\begin{algorithm}
		\SetKwFunction{aabbtree}{computeAABBTree}
		\SetKwFunction{computeRay}{computeRay}
		\SetKwFunction{computeDistance}{computeDistance}
		\SetKwFunction{checkIntersection}{checkIntersection}
		\SetKwInOut{KwIn}{Input}
		\SetKwInOut{KwOut}{Output}
		\KwIn{$P(\mathbf{q})$ and $M(\mathbf{q})$ for $l = \lbrace 1, 2, \dots, L \rbrace$} 
		\KwOut{$\hat{P}$}
				
		$\hat{P} = [\ ]$
		
		$\mathcal{T} = \aabbtree(\mathcal{M}(\mathbf{q}))$
		
		\For{$i \leftarrow 1$ \KwTo $L$}{
				\ForEach{$p \in P_j$}{
					$\mathbf{r} = \computeRay(\mathbf{p},x_j)$
					
					$\mathbf{p_{\cap}} = \checkIntersection(\mathcal{T},\mathbf{r})$
					
					\If{$\mathbf{p_{\cap}} == NULL$ \textbf{or} $\computeDistance(\mathbf{p},\mathbf{p_{\cup}}) > \epsilon$}
					{
						$\hat{P} = \hat{P} \cup \mathbf{p}$
					}
				}
		}
		\KwRet{$\hat{P}$}
		\label{alg:intersection}
		\caption{Robot Model Removal}
	\end{algorithm}

Alg. 1 is executed at each time instant, for each number of links $l$. In terms of time complexity, it requires the construction of the AABB tree from the mesh which is $V\log(V)$, where $V$ is the number of primitives to be inserted into the tree. The costs for the intersection check for the single sensor measurement is logarithmic with respect to the number of faces in the mesh, while the distance computation between $\mathbf{p}$ and $\mathbf{p_{\cup}}$ is constant. Therefore the overall cost is given by: 
\begin{equation}
	\underbrace{O(V\log(V))}_{\textit{Tree construction}}  + 
	\underbrace{ O(NM\log(F)) }_{\textit{Intersection check}} +
	\underbrace{O(NM)}_{\textit{Distance computation}}
\end{equation}
 
However, in our condition the point cloud generated from the proximity sensor has small resolution and the number of sensors embedded into the robot body is limited. Therefore, $V \gg MN$ with the overall computational complexity leading to $V\log(V)$.

\subsection{Minimum Distance Points}
Once the robot's model is filtered out, the resulting point cloud only provides information about the external environment. After performing a simple 3D spatial filtering to remove static elements such as tables and parts of the room geometry, we are only left with the measurements corresponding to the obstacles in proximity to the robot body. The proposed algorithm, for each AABB tree of the set $\mathcal{AT}$ computed for the robot links , iterates over the entire point cloud of the environment $\hat{P}$ , finds the pair of closest points respectively on the robot model ${mpr}_l$ and on the filtered environment cloud  $Amp_l$, compute their square minimum distance, and store the collected information into the array $ASQ$. \\ Finally, to find the absolute minimum value, the function findMinimumLink is called, which identifies the link and the pair of closest points that will be used by the reactive controller to avoid the identified obstacles.\\
The full algorithm is showed in Alg. 2.
\begin{algorithm}

    \SetKwFunction{findminimum}{findMinimumLink}
    \SetKwFunction{square}{squaredistance}
    \SetKwFunction{closest}{closestPoint}
    \SetKwInOut{KwIn}{Input}
	\SetKwInOut{KwOut}{Output}
    \KwIn{$\hat{P}$, $\mathcal{AT}$}
    \KwOut{$ASQ_l$, $mpr$, $Amp_l$, $l$} 

    $ASQ=[\ ]$
    
    $Amp_{\hat{P}} =[\ ] $
    
    \For{$i \leftarrow 0$ \KwTo $\mathcal{AT}$ $size$}{
        \ForEach{$p \in \hat{P}$}{
            $ASQ_i$=$\mathcal{AT}_i\square($p$)$
                
            \If{$ASQ_i < ASQ_{i-1}$} 
            {
                $\mathbf{cmd}=ASQ_i$
                
                $Amp_i=p$
            }
        }
    }
    $\mathbf{l}=\findminimum(ASQ)$
    
    $\mathbf{mpr}=\mathcal{AT}_l\closest(Amp_l)$
    
    \KwRet{$ASQ_l$, $mpr$, $Amp_l$, $l$}
    \label{alg:mindist}
    \caption{Find minimum point}
\end{algorithm}

In terms of time complexity, it requires the execution of the outermost loop, which is $M$, by the execution of the inner loop, which depends on the size of $\hat{P}$, that is N, and by the search within the AABB trees, which is $\log(k)$, where k is the number of primitives that make up the tree. In addition it should necessary to consider the minimum search function, but since the array is composed of 6 elements, it can be neglected.\\ Hence the actual cost is:
\begin{equation}
	\underbrace{O(MN\log(K))}_{\textit{Minimum square distance searching and iterations}} 
\end{equation}
The outcomes of the cloud processing pipeline are visualized in multiple configurations in Fig. \ref{fig:mindist_methods}.

\subsection{Motion Control Algorithm}
The control scheme implemented for this activity is based on the "Task Priority Control framework" presented in \cite{Simetti}, that allows to satisfy several control objectives with different priorities. Specifically, in our application we consider a robot controller with two priority layers. The mathematical proof and formalism for the task priority controller can be found in \cite{albiniEnablingNaturalHumanrobot2017}.

\subsubsection{Safety control objective}
The highest priority controller deals with the safety of the overall system, implementing the obstacle avoidance behaviour based on the pre-processed proximity sensors feedback. The goal is to increase the distance from the identified obstacles while trying to satisfy action-defining lower priority control objectives, such as reaching a desired goal position with the end-effector. 

By defining \(\dot{{x}}_{o}\) the velocity estimated for the obstacle closest to the robot body, \(d_{min} \in \Re\) the minimum distance that the robot should keep from the detected obstacle, \(\Delta\) the constant offset used to create a buffer zone to smoothly activate the safety control task and \(d_{curr} \in \Re\) the current minimum distance computed with \ref{alg:mindist}, we can write the feedback reference rate as:

\begin{equation}
    \dot{\bar{x}} \overset{\Delta}{=} \dot{{x}}_{o} + \lambda(d_{min} + \Delta - d_{curr}), \ \ \ \lambda > 0
\end{equation}

The sigmoid activation function regulating the activity of the safety control task is described by: 

\begin{equation}
        a^i(x) = 
        \begin{cases}
            1  & \mbox{if } x < d_{min} \\
            s(x)  & \mbox{if } d_{min} < x < d_{min} + \Delta \\
            0  & \mbox{if } x > d_{min} + \Delta
        \end{cases}
\end{equation}

The task induced Jacobian $^{0}J_{mp/0}$, that is the Jacobian of minimum point distance wrt the base, and is obtain as the folllowing:

\begin{equation}
     ^{0}J_{mp/0}=^{0}S_{mp/lmp} \;^{0}J_{lmp/0}
 \end{equation}
 
where $S$ is the \textbf{rigid body Jacobian} defined as following:

\begin{equation}
     ^{0}S_{mp/lmp}  \overset{\Delta}{=} 
     \begin{bmatrix}
    \mathbb{I}_{3 \times 3} & [^{0}r_{mp/lmp}]^\top\\
    0_{3 \times 3} & \mathbb{I}_{3 \times 3}
     \end{bmatrix} \in \Re^{6x6}
 \end{equation}
 such that $[^{0}r_{mp/lmp}]^\top$ is the skew symmetric distance between the point of minimum distance $mp$ and the frame of the link $lmp$ to which the point belongs.
\subsubsection{Goal-reaching control objective}
The goal reaching task is assigned the lowest priority in the proposed two-layer control architecture and is executed within the null space of the primary, safety-related task. Specifically, this action-defining, equality control objective is used to reduce reduce the position and orientation error between the robot end-effector and the desired goal frame. 
The feedback reference rate for the end-effector position control was computed as:

\begin{equation}
    \dot{\bar{\textbf{x}}} \overset{\Delta}{=} \lambda(\textbf{x}_{goal} - \textbf{x}_{e\_e}) \ , \lambda > 0
\end{equation}

where \(\textbf{x}_{goal} \in R ^{3x1}\) is the position of the goal, while \(\textbf{x}_{e\_e} \in R ^{3x1}\) is the current tool position, both computed with respect to the robot base.

The feedback reference rate for the end-effector orientation control was computed as:

\begin{equation}
    \dot{\bar{\textbf{x}}} \overset{\Delta}{=} \lambda\textbf{n}\theta , \ \ \ \lambda > 0
\end{equation}

where \(\textbf{n}\theta \in \Re^{3x1}\) is the angle-axis representation of the orientation error.

The task-induced Jacobian for the goal reaching task was simply computed considering the linear and angular components of the robot end-effector Jacobian matrix:

\begin{equation}
    ^{0}J_{e/0}  \overset{\Delta}{=} \begin{bmatrix}
           ^{0}J_{e/0}^L \\\
           ^{0}J_{e/0}^A \\
\end{bmatrix} \in \Re^{6x6}
\end{equation}

The obtained task vectors and Jacobians are then exploited to compute suitable joint velocity references according with the control formulation described in \cite{albiniExploitingDistributedTactile2021}.

%% file: 03_Experimental_Validation/experiments.tex

\begin{figure*}[t]
	\centering
	\subfigure[]{\includegraphics[scale=0.5]{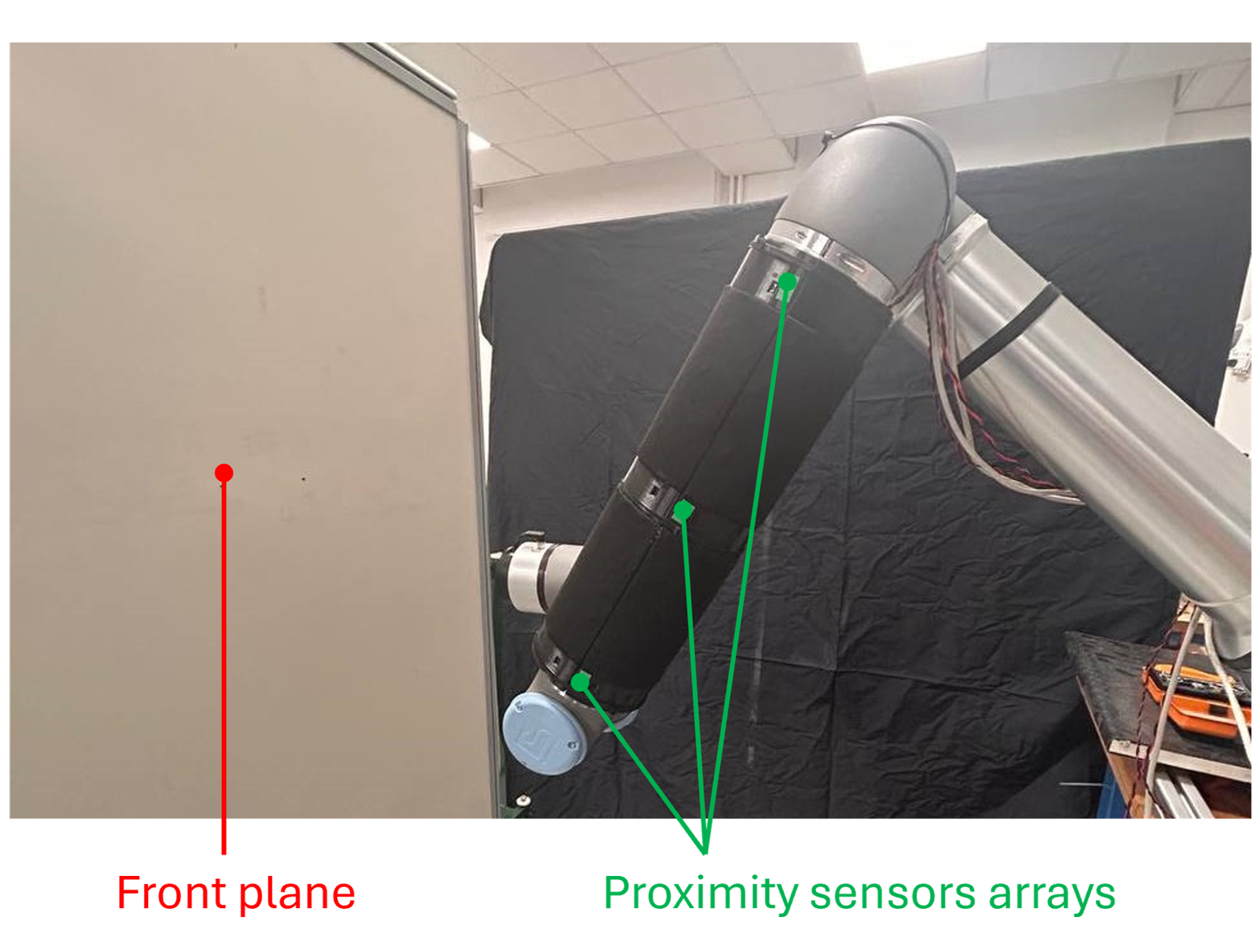}}
        \subfigure[]{\includegraphics[scale=0.5]{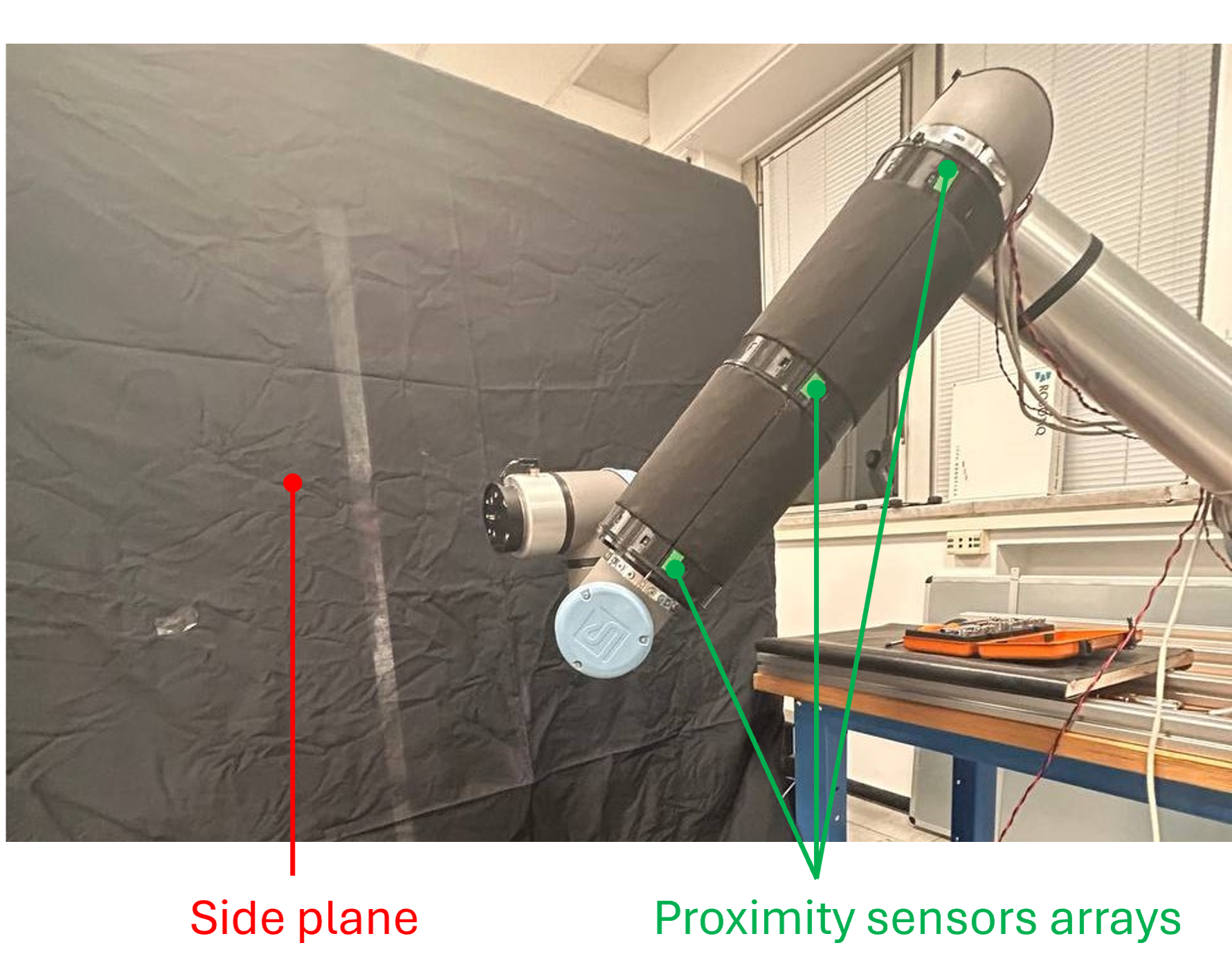}}
        \subfigure[]{\includegraphics[scale=0.5]{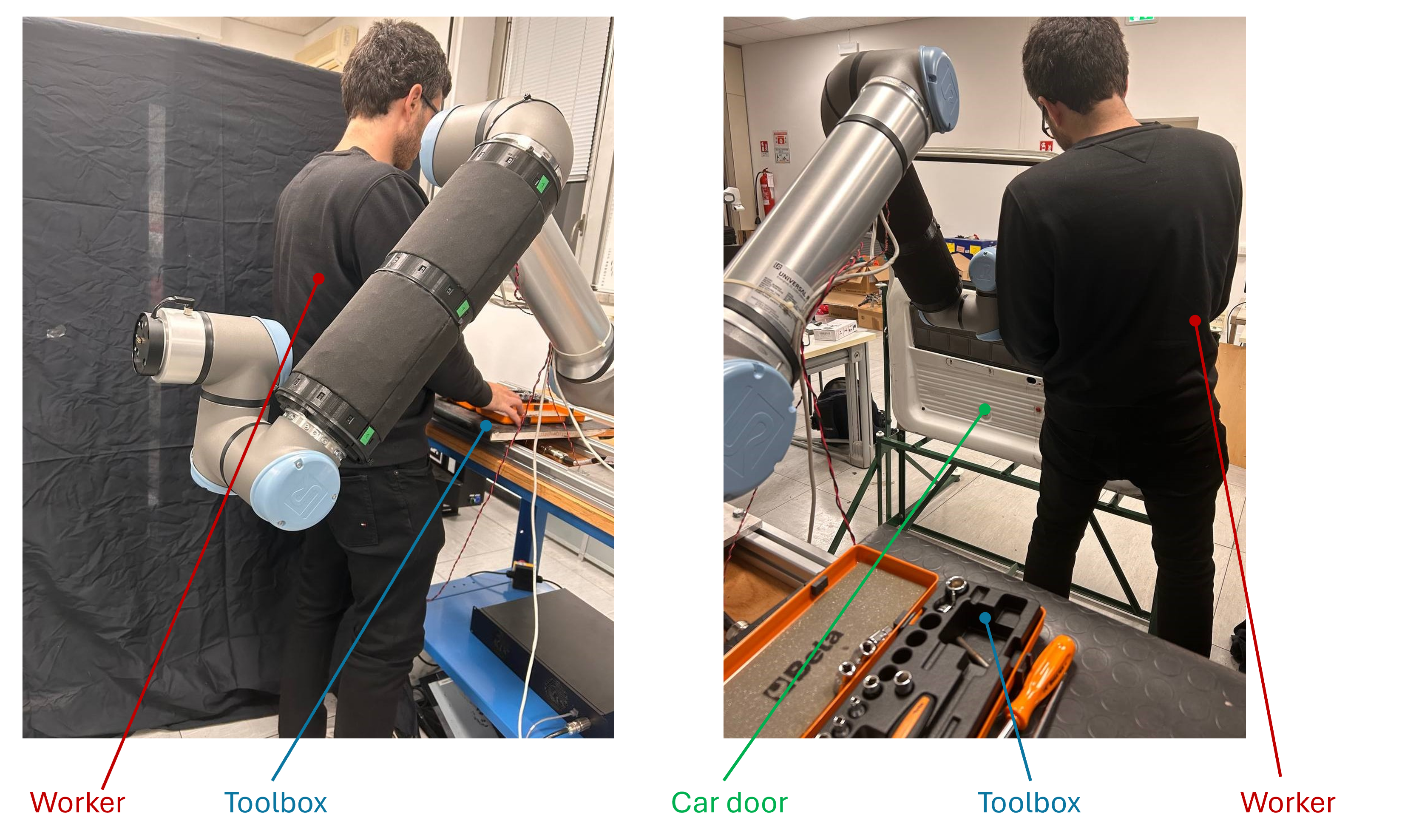}}
	\caption{Experimental setup configurations. (a) A static planar obstacle is positioned in front of the robot, to evaluate algorithm's behavior on the uncovered flange. (b) Another planar object is placed besides the robot, to evaluate the algorithm's behavior on the uncovered links. (c) Human-robot Collaboration mockup. A toolbox is placed on the left or right of the robot base depending on the trial configuration. A car door is positioned in front of the robot's workspace, being the target for the user's actions.}
	\label{fig:experimental_setup}
\end{figure*}

\label{sec:experiments}
The proposed approach was validated using a 6-degree-of-freedom (6-DoF) UR10e robotic arm by Universal Robotics, which was partially equipped with multi-zone time-of-flight (ToF) sensors. Specifically, VL53L8CX sensors, manufactured by ST Microelectronics \cite{vl53l8cx_documentation}, were integrated into flexible PCB arrays as part of the ProxySKIN technology\cite{Giovinazzo}. Each proximity sensor generates an 8-by-8 matrix of distance measurements and is characterized by a field of view of \(45^\circ\). 
In this setup, three ring-shaped arrays, each consisting of 10 sensors, were mounted around the third link of the UR10e robot using custom 3D printed covers, as shown in Figure \ref{fig:experimental_setup}. The rest of the robot body was not sensorized to test the effectiveness of our algorithm.

The first set of experiments was conducted in a static environment, using fixed obstacles in proximity to the robotic platform. Specifically, in a first trial we placed a planar surface in front of the robot body (Figure \ref{fig:experimental_setup}-a), while in a second study we placed it on right side of the UR10e (Figure \ref{fig:experimental_setup}-b). In both cases, we defined a goal frame for the robot's end-effector beyond the planar surface and, by activating the obstacle avoidance safety control task, we checked the robot response to two different types of controllers: 
\begin{itemize}
    \item A state of the art reactive controller, using the proximity sensors positions on the robot body as control points;
    \item The controller explained in section \ref{sec:methodology}, using any point on the robot model -- including non-sensorized links --  to enable reactive motion behaviors. 
\end{itemize}

In the second validation scenario, the robot was operating within a small collaborative workspace alongside a human operator. 

As illustrated in Figure \ref{fig:experimental_setup}-c, the robot was positioned on a table, and a car door served as a mockup for screwing tasks. Specifically, the robot executed a motion task following a set of waypoints to simulate screwing operations on the car door. The actual screwing process was omitted as it is not relevant to this study.

The operator’s task involved moving back and forth the table and the car door to simulate retrieving tools and screws from a toolbox and positioning them in the car door holes, awaiting the robot to perform screwing. Notably, the operator didn't follow specific rules but acted independently of the robot’s movements. Additionally, the toolbox was positioned near the robot’s base, requiring the operator to approach the first and second links, which were lack sensor coverage. This setup is particularly useful for testing avoidance behaviors on non-sensorized parts of the robot.

To validate our approach, we compare its performance to the method described in \cite{caroleo2024}, where only the mounting points of the sensors are used to move the robot away from obstacles. We evaluate the performance of both algorithms by measuring the distance between the robot’s point cloud and the closest point of the environment’s point cloud over time.

We conduct ten trials for each algorithm. Specifically, during the first five trials, the toolbox is placed on the left side of the robot, while in the remaining five trials, it is positioned on the right. This placement variation allows to observe the robot's performance under differing environmental conditions.

\begin{figure}[t]
    \centering
    \includegraphics[width=0.5\textwidth]{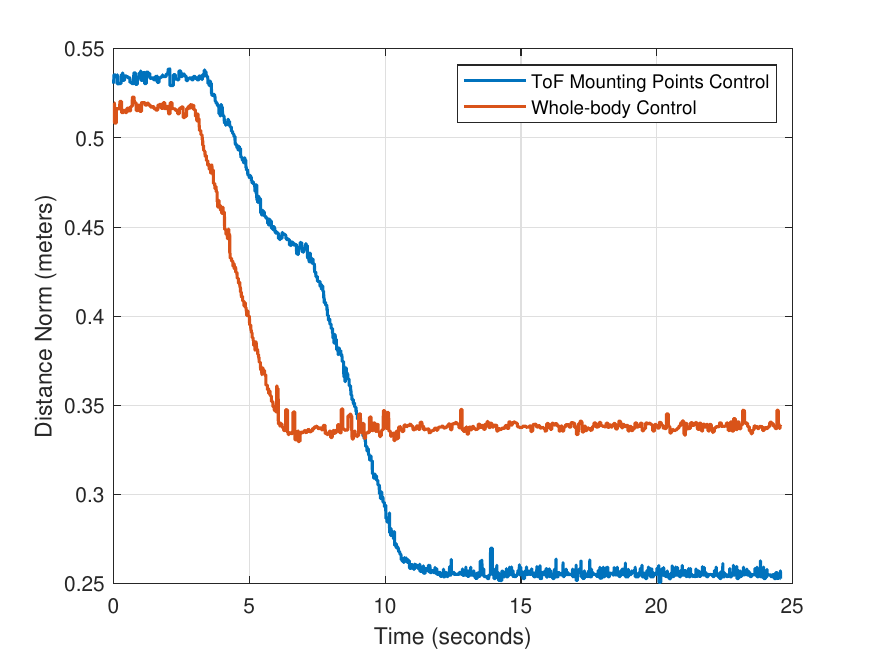}
    \caption{Comparative analysis between the presented whole-body algorithm and the baseline during the front plane test.}
    \label{fig:exp_front_wall}
\end{figure}

%% file: 04_Results_and_Discussion/discussion.tex

\label{sec:discussion}
In this section we present quantitative results obtained from our experimental trials, providing a comparison between our proposed whole-body algorithm \textit{\textbf{(WB)}} and the baseline state-of-the-art approach based on motion control of the sensors' mounting points \textit{\textbf{(SM)}}.

\subsection{Static Obstacles Interaction}
\begin{figure}[t]
    \centering
    \includegraphics[width=0.5\textwidth]{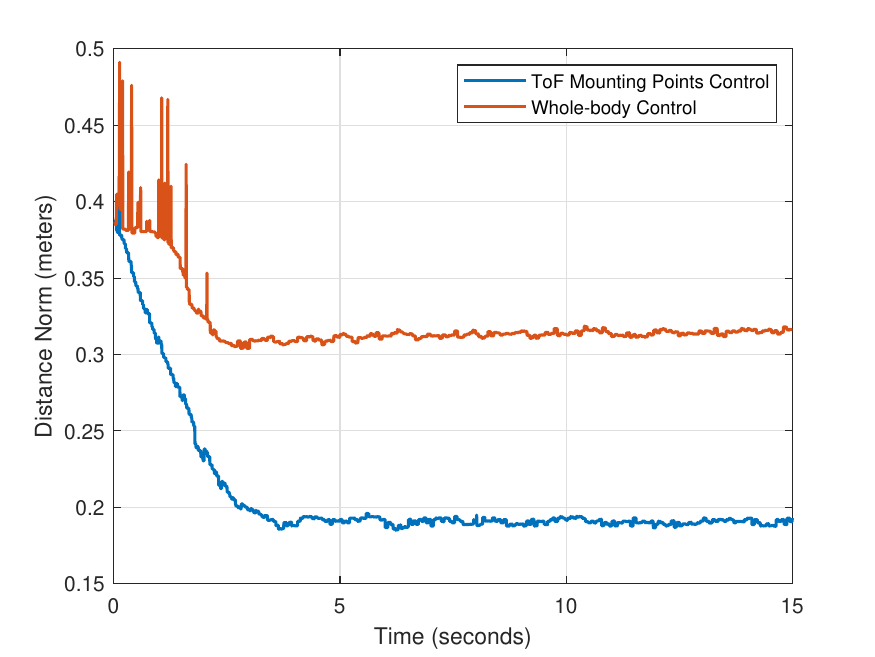}
    \caption{Comparative analysis between the presented whole-body algorithm and the baseline during the side plane test.}
    \label{fig:exp_side_wall}
\end{figure}

We have evaluated the minimum distance between the robot and the environment point cloud while performing a 6-dof reaching task in the presence of a static planar obstacle. Trials were performed with the obstacle being placed both in front and on the left side of the robot's workstation. The robot attempts to reach a cartesian goal located directly on the obstacle's surface, therefore this set of experiments aims to show how the proposed algorithm helps the robot to keep the distance with unmodeled objects. In both situations we have evaluated the norm of the minimum distance between the point clouds representing the environment and the robot, respectively. 
The outcome of the trial performed with the front panel is shown in Figure \ref{fig:exp_front_wall}: it can be easily noticed that when the whole-body algorithm is used the robot is able to keep a greater distance with respect to the \textit{\textbf{(SM)}} counterpart. The robot exhibits this behavior because with the \textit{\textbf{(WB)}} algorithm the control point is located on the flange, that is in fact the link closest to the obstacle.
A similar result was obtained considering a static obstacle on the left side of the robot's workstation, whose outcomes are shown in Figure \ref{fig:exp_side_wall}. As for the previous case, our novel approach allows the robot to keep a larger safety distance between external objects. In this case, the robot's end-effector is moved sideways, therefore the control point is located on the side of the non-sensorized link.

\begin{figure*}[t]
	\centering
	\subfigure[]{\includegraphics[scale=0.5]{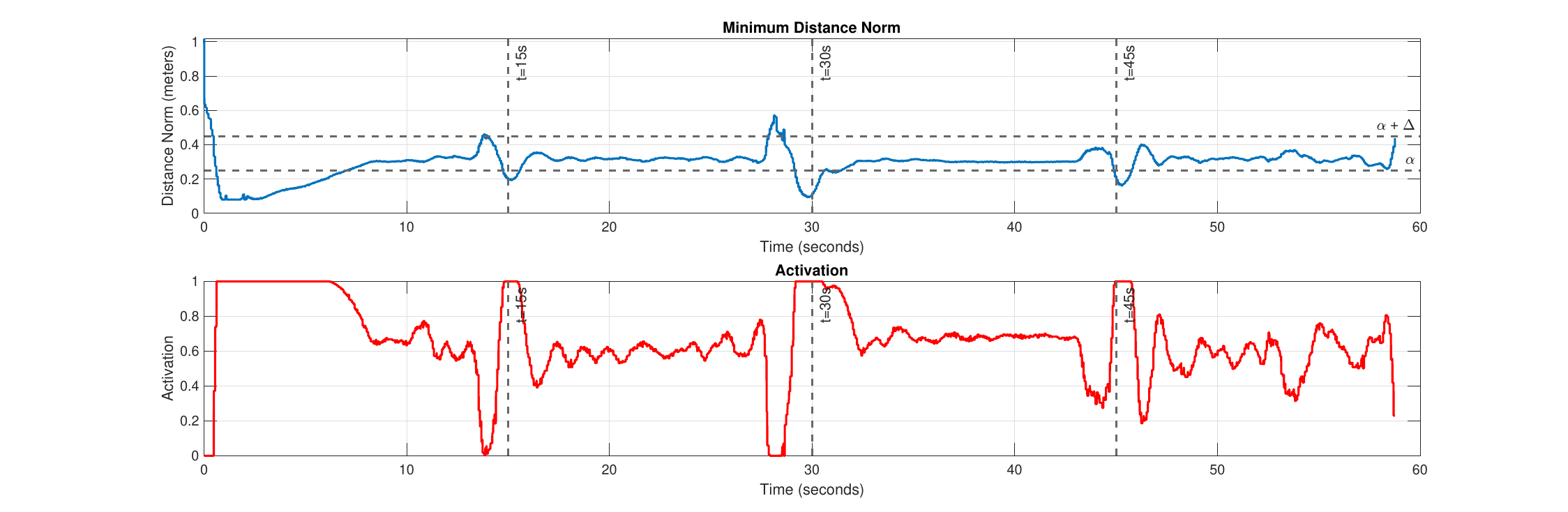}}
        \subfigure[]{\includegraphics[scale=0.5]{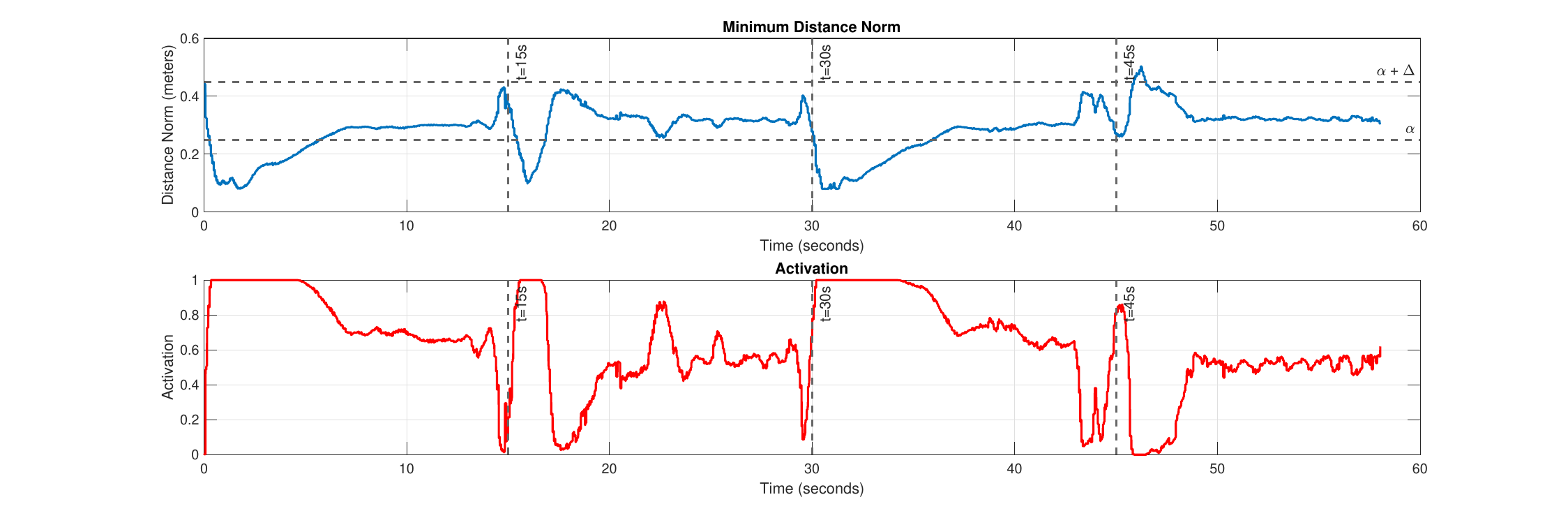}}
	\caption{Time-based plots of two complete hri trials. (a) shows data related to a trial performed with the baseline algorithm, with the toolbox placed to the right of the robot. (b) shows the outcome of a trial performed with the novel whole-body approach. The toolbox was placed on the left side instead.}
	\label{fig:hri_trials}
\end{figure*}

\subsection{HRI in a Collaborative Scenario}

Along with a proof-of-concept evaluation on static scenarios with unmodeled obstacles, we evaluated the proposed algorithm on a HRI task, to show how it can contribute to enhance safety whenever an operator works within the robot's workspace. We have defined four operational conditions to be performed within the mockup collaborative cell shown in Fig. \ref{fig:experimental_setup}:
\begin{itemize}
\item \textit{\textbf{SR}}: The operator reaches the right \textit{\textbf{(R)}} side of the robot's mounting table and moves towards the car door by keeping the right side of the workcell. The robot's avoidance behavior is controlled through the sensor's mounting points.
\item \textit{\textbf{SL}}: The operator reaches the left \textit{\textbf{(L)}} side of the robot's mounting table and moves towards the car door by keeping the left side of the workcell. The robot's avoidance behavior is controlled through the sensor's mounting points.
\item \textit{\textbf{WBR}}: The operator reaches the right \textit{\textbf{(R)}} side of the robot's mounting table and moves towards the car door by keeping the right side of the workcell. The robot's avoidance behavior is controlled through the whole-body avoidance algorithm.
\item \textit{\textbf{WBL}}: The operator reaches the left \textit{\textbf{(L)}} side of the robot's mounting table and moves towards the car door by keeping the left side of the workcell. The robot's avoidance behavior is controlled through the whole-body avoidance algorithm.
\end{itemize}

\begin{table}[t]
\centering
\caption{Measured distances and their corresponding maximum standard deviations for the selected operational conditions.}
\label{tab:hri_analysis}
\setlength{\tabcolsep}{10pt} 
\renewcommand{\arraystretch}{1.5} 
\begin{tabular}{l|ll}
    & $\overline{d}$(m) & $\sigma_{max}$(m) \\ \hline
SR  & 0.281    & 0.135   \\
SL  & 0.169    & 0.100   \\
WBR & 0.279    & 0.097   \\
WBL & 0.295    & 0.214   \\
\end{tabular}
\end{table}

We have observed the robot's behavior over 40 trials, lasting approximately 1 minute each. In particular, 10 trials for each of the four situations were performed. The two algorithms were compared by observing the minimum distance between the robot and the environment, both represented as point clouds. In particular, we have evaluated the average value of the minimum distance $\overline{d}$ and the maximum value of its standard deviation $\sigma_{max}$ over the 10 trials of each scenario. The results have meed summarized in Table \ref{tab:hri_analysis}.

The analysis on the average minimum distance shows that \textit{\textbf{WB}} allows the robot to keep a greater minimum distance regardless of the operator's position. In particular, given the specific configuration we have used for our setup, when the operator is moving within the left side of the cell, the sensors' mounting locations are farther than the non-sensorized links, explaining why the \textit{\textbf{WB}} trials show the worst performance. In contrast, in both \textit{\textbf{WBL}} and \textit{\textbf{WBR}} the robots keeps a minimum distance of almost 30 cm.   
Moreover, $\sigma_{max}$ values provide an insight on the responsiveness of the system, since higher values reasonably correspond to wider avoidance movements. In particular, \textit{\textbf{WBL}} trials exhibit a noticeably larger maximum value, mostly due to the operator being in the proximity of two different non-sensorized links (i.e. the tool and the first link).

Figure \ref{fig:hri_trials} shows the minimum distance and the avoidance activation function over two trials performed in \textit{\textbf{SL}} and \textit{\textbf{WBR}} conditions respectively. Vertical markers indicate the time points at which the operator moves within the workspace. Horizontal lines in the minimum distance plots represent the thresholds that define the transition region of the activation function. It's clearly visible how the operator's movements cause a sudden increase in the activation in both experiments. 
Figure \ref{fig:hri_trials} also shows that our proposed method shows comparable performance with a state-of-the-art algorithm that was validated in numerous interaction scenarios.

%% file: 05_Conclusions/conclusions.tex
\label{sec:conclusions}
In this paper we presented a novel proximity servoing algorithm that generates reactive collision avoidance behavior by exploiting a limited set of multi-zone range sensors integrated on a robotic manipulator. In particular, we have proposed a methodology that leverages the geometric model of the robot to find its closest point with respect to the environment, represented as a point cloud. We have compared our method with a well-established sensor-based algorithm for reactive avoidance, showing comparable qualitative and quantitative performances. Moreover, the presented algorithm can exploit any arbitrary point on the robot's surface to perform avoidance motion, showing improvements in the distance margin due to the rendering of a virtual avoidance behavior on non-sensorized links. To the best of our knowledge, this is the first paper addressing whole-body obstacle avoidance with partial robot sensorization. Our approach can also represent a neat improvement in terms of integration efforts, since it allows to endow a whole robot with proximity avoidance without the need of covering the whole surface with sensors or having external sensing architectures such as cameras.
In future works, we plan to perform extensive trials in hri scenarios with multiple subjects, as well as a coverage analysis to cope with blind spots that could arise due to limited sensorization.